\documentclass{article}
\usepackage{ijcai20}

\usepackage{times}

\usepackage{soul}
\usepackage{url}
\usepackage[utf8]{inputenc}
\usepackage[small]{caption}
\usepackage[pdftex]{graphicx}
\usepackage{amsmath,amssymb}
\usepackage{booktabs}
\usepackage{algorithm}
\usepackage{algpseudocode}

\title{Discriminator Soft Actor Critic without Extrinsic Rewards}

\author{
Daichi Nishio$^1$\and
Daiki Kuyoshi$^1$\and
Toi Tsuneda$^1$\footnote{Contact Author}\And
Satoshi Yamane$^1$
\\
\affiliations
$^1$Kanazawa University\\
\emails
\{dnishio, dkuyoshi, ttsuneda\}@csl.ec.t.kanazawa-u.ac.jp, 
syamane@is.t.kanazawa-u.ac.jp 
}

\begin{document}

\maketitle

\begin{abstract}
It is difficult to be able to imitate well in unknown states from a small amount of expert data and sampling data. 
Supervised learning methods such as Behavioral Cloning do not require sampling data, but usually suffer from distribution shift. 
The methods based on reinforcement learning, such as inverse reinforcement learning and generative adversarial imitation learning (GAIL), can learn from only a few expert data. 
However, they often need to interact with the environment. 
Soft Q imitation learning addressed the problems, and it was shown that it could learn efficiently by combining Behavioral Cloning and soft Q-learning with constant rewards. 
In order to make this algorithm more robust to distribution shift, we propose Discriminator Soft Actor Critic (DSAC). 
It uses a reward function based on adversarial inverse reinforcement learning instead of constant rewards. 
We evaluated it on PyBullet environments with only four expert trajectories.
\end{abstract}

\section{Introduction}

Recent developments in the field of deep reinforcement learning have made it possible to learn diverse behaviors for high-dimensional input. 
However, there are still some problems. Among them, we focus on the efficiency of learning and the difficulty of designing a reward function.
For example, when using reinforcement learning for artificial intelligence of autonomous driving, it is necessary to deal with many unexpected phenomena such as various terrain and people coming out.
If we design a reward function to solve this problem, the program may become enormous. 
Also, incomplete reward function design may promote unexpected behavior.
In addition, it is necessary to explore with many random actions until the agent obtains a reward in an environment with sparse rewards.

In such setting of problems, imitation learning is often used instead of reinforcement learning.
Behavioral Cloning~\cite{pomerleau1991efficient}, which is the classical imitation learning, is a simple supervised learning algorithm that maximizes the likelihood of the actions taken by an expert in a certain state. 
It shows good results for simple tasks, but it requires a large dataset of the pairs of state and action, and it sometimes behaves strangely in a state is not in the dataset.
In order to overcome these disadvantages, inverse reinforcement learning performs a two-step learning process in which it estimates a reward function instead of expert actions, and it performs reinforcement learning based on the reward function. 
This algorithm helps to learn to behave in unexpected situations.

However, inverse reinforcement learning has the disadvantage of being unstable due to two stages of learning.
Therefore, the methods of learning the behavior of the expert directly by generative adversarial learning without explicitly finding a reward function were proposed such as Generative Adversarial imitation Learning (GAIL)~\cite{ho2016generative}. 
They make it possible to learn efficiently even with a small amount of data, and research based on it is being researched even now.

Although Behavioral Cloning was no longer considered useful, Reddy et al.~\shortcite{reddy2019sqil} proposed Soft Q Imitation Learning (SQIL), which addressed state distribution shift by the combination of Behavioral Cloning and reinforcement learning.
It has been reported that the learning has been performed efficiently with less training steps than in previous adversarial imitation learning.
In this paper, we point out the improvements of this method and propose more efficient and more robust algorithm.
We evaluate it with four environments of PyBullet~\cite{coumans2016pybullet} and we show the strong and weak points.

\section{Background}
We consider problems that satisfy the definition of a Markov Decision Process (MDP).
In continuing tasks, the returns for a trajectory $ \tau = {(s_t, a_t)}^{\infty}_{t=0} $ are defined as $ r_t = \sum ^{\infty}_{k=t} \gamma ^{k-t} R(s_k, a_k) $, where $ \gamma $ is a discount factor.
In order to use the same notation for episodic tasks, we can define a set of absorbing state $ s_a $. When we define the reward $ R(s_a, \cdot) = 0$, we can define returns simply as $ r_t = \sum ^{T}_{k=t} \gamma ^{k-t} R(s_k, a_k) $.  In reinforcement learning, we would like to learn a policy $ \pi $ that maximizes expected returns.
Recently, maximum entropy reinforcement learning has given an outstanding performance especially in a complex environment. 
It provides a substantial improvement of an exploration and a robustness.
Soft Actor Critic (SAC)~\cite{haarnoja2018soft}, which optimizes a stochastic policy in an off-policy way, is one of them, so this algorithm has a good sample efficiency. The maximum entropy objective generalizes the standard objective by augmenting it with an entropy term $ \mathcal{H} $. Optimal policy additionally aims to maximize its entropy,
\begin{equation}
 \mathrm{arg}\max_{\pi} \sum ^{\infty}_{t=0} \sum ^{\infty}_{k=t} \gamma ^{k-t} \mathbb{E} \left[ R(s_k, a_k) + \alpha \mathcal{H} (\pi (\cdot | s_k)) | s_k, a_k \right].
\end{equation}
where $ \alpha $ is the temperature parameter that determines the relative importance of the entropy term and the reward. 
In~\cite{haarnoja2018soft2}, the optimal dual variable $ \alpha _ t $ is solved after solving the optimal action value $ Q ^ *$ and policy $ \pi ^* $ as Equation~\ref{min_temp}.
\begin{equation}
    \mathrm{arg} \min_{\alpha_t} \mathbb{E} \left[ - \alpha _ t \log \pi ^ {*} _ {t} (a_t|s_t; \alpha_t) - \alpha _ t \tilde{\mathcal{H}} \right]
    \label{min_temp}
\end{equation}

\section{Related work}

Behavioral Cloning~\cite{pomerleau1991efficient} is a classical imitation learning algorithm.
It is a method of supervised learning that takes the state $ s _ E $ in the expert data as input and regards the action $ a _ E $ of the expert as the label.
Assuming that the transition information about the expert is $ \tau = (s_0, a_0, s_1, a_1, ..., s_T) $, the objective function for the parameter $ \theta $ has a form that maximizes the log likelihood as follows. 
\begin{equation}
    \label{eq_BC}
    \max _ \theta \sum^{T - 1}_{t = 0} \log \pi_\theta (a_t | s_t)
\end{equation}
Although this method is simple, it is easy to overfit to the expert data because it does not learn the result of the action, and it suffers from the state distribution shift. 
As a result, it has the disadvantage of not being able to make good decisions for unseen states. 

In the practical problem, there is a limit on simple supervised learning like Behavioral Cloning. 
Therefore, we consider an approach of learning in the framework of reinforcement learning. 
For that purpose, it is necessary to define a reward function $ R(s, a)$. Inverse Reinforcement Learning (IRL) is one of the ways. 
There are two major problems in applying it as design a reward function. 
One is that there can be multiple reward functions that lead to the optimal policy. 
This makes it difficult to determine the reward function and makes the learning unstable.
The other is that the state space becomes huge for large-scale problems, and the constraints become too much to obtain a feasible solution.

Ho and Ermon~\shortcite{ho2016generative} proposed Generative Adversarial Imitation Learning (GAIL), which is an imitation learning of nonlinear cost function using GANs~\cite{goodfellow2014generative}.
This makes it possible to perform model-free learning, which requires inverse reinforcement learning and requires less expert data.
The objective function is defined as follows,
\begin{equation}
    \max _ {D} \mathbb{E}_{\pi}  \left[\log (D(s, a))\right] + \mathbb{E}_{\pi_{E}}[\log (1-D(s, a))]-\lambda \mathcal{H} (\pi),
\end{equation}
where $ D $ is a discriminative classifier to distinguish between the distribution of data generated by $ \pi_{E} $ and the true data distribution. 
GAIL has become the current mainstream imitation learning method and is available for many tasks, however, it has the disadvantage of requiring many samples from environments.

Finn et al.~\shortcite{finn2016connection} shows that in contrast with energy based inverse reinforcement learning can be solved using Guided Cost Learning (GCL), it can be formulated in the same way as GANs. 
In this algorithm, the reward function $ R $ is defined as follows by trajectory $ \tau $,
\begin{equation}
    R(\tau)=\log \left(D_{\theta}(\tau)\right)-\log \left(1-D_{\theta}(\tau)\right),
\end{equation}
where $ D _ {\theta} (\tau) $ is a discriminative classifier parameterized by $ \theta $.

Adversarial Inverse Reinforcement Learning (AIRL)~\cite{fu2017learning} is based on this idea.
It is transformed into a reward function that depends only on the current state $ s $ and action $ a $ so that it can be more robust.

Kostrikov et al.~\shortcite{kostrikov2018discriminator} proposed Discriminator Actor Critic (DAC). They focused on the bias of rewards, which is a problem for GAIL and AIRL. 
They pointed out that some tasks may not converge to the optimal policy because the methods implicitly determined the reward of the terminal state to be zero.
In order to design a reward function without bias, DAC also learns the terminal state. 
In addition, GAIL uses the entire trajectory, whereas the sample efficiency is improved by using the experiences stored in the replay buffer $ \mathcal{B} $ during off-policy training like AIRL.
At this time, It is necessary to do importance sampling.
\begin{eqnarray}
    \max _{D} & \mathbb{E}_ {\mathcal{B}} & \left[\frac{p_{\pi_{\theta}}(s, a)}{p_{\mathcal{B}}(s, a)} \log (D(s, a))\right] \nonumber \\ &+ & \mathbb{E}_{\pi_{E}}[\log (1-D(s, a))]-\lambda \mathcal{H} (\pi)
\end{eqnarray}
However, it is practical to train without importance sampling because it is difficult to converge.

\begin{figure*}[htbp]
    \centering
        \includegraphics[width=0.9\hsize]{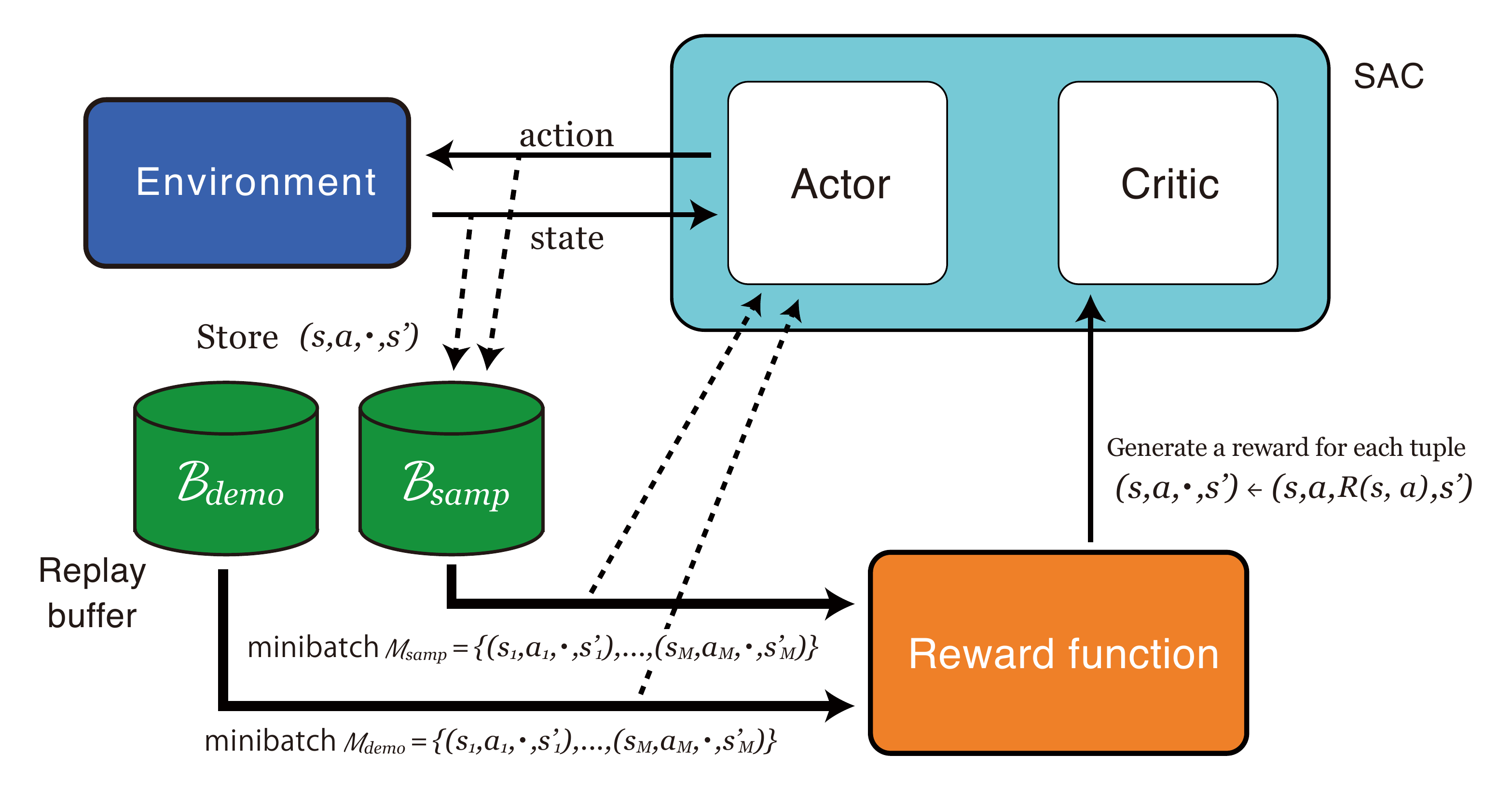}
    \caption{The overall of DSAC algorithm.}
    \label{fig:overall}
\end{figure*}

While most of these researches were derived from GAIL, Reddy et al.~\shortcite{reddy2019sqil} proposed Soft Q Imitation Learning (SQIL), which addressed state distribution shift by the combination of Behavioral Cloning and reinforcement learning.
In this method, even in the environment with sparse rewards it can be changed to Q-learning or off-policy actor critic with a small change of the code.
In the case of soft Q-learning, it is known that the optimal policy $ \pi (a | s) $ is as follows,
\begin{equation}
    \pi (a | s) = \exp \left(\frac{Q(s,a) - V(s)}{\alpha} \right),
\end{equation}
where $ \alpha $ is the temperature parameter. 
If we assume that the behavior of our agent follows the policy, we can define the loss function by Equation \ref{eq_BC}.
\begin{equation}
    \label{lossBC}
    \ell_{\mathrm{BC}}(\theta) \triangleq \sum_{(s, a) \in \mathcal{B}_{\mathrm{demo}}}-\left(\frac{Q_{\theta}(s, a) - V_{\theta}(s)}{\alpha}\right)
\end{equation}
SQIL aims to learn by considering the trajectory by regularizing this loss function with squared soft Bellman error $ \delta^{2}(\mathcal{B}, r)$, called Regularized Behavioral Cloning (RBC).
\begin{equation}
    \ell_{\mathrm{RBC}}(\theta) \triangleq \ell_{\mathrm{BC}}(\theta)+\lambda \delta^{2}\left(\mathcal{B}_{\mathrm{demo}} \cup \mathcal{B}_{\mathrm{samp}}, 0\right) \label{RBC}
\end{equation}
\begin{equation}
    \delta^{2}(\mathcal{B}, r) \triangleq \frac{1}{|\mathcal{B}|} \sum_{\left(s, a, s^{\prime}\right) \in \mathcal{B}}\left(Q_{\theta}\left(s, a\right)-\left(r+\gamma V_{\theta}\left(s\right) \right)\right)^{2},
    \label{softBellman}
\end{equation}
where $ \lambda \in \mathbb{R}_{\geq 0} $ is a hyperparameter that determines the relative importance of Behavioral Cloning
versus soft Q-learning. In addition, $ \mathcal{B}_{\mathrm{demo}} $ and $ \mathcal{B}_{\mathrm{samp}} $ are a replay buffer of demonstration data by an expert and sampling data from an environment by an agent respectively.

Furthermore, we can rewrite the gradient of $ \ell_{\mathrm{RBC}}(\theta) $ as simple form.
\begin{eqnarray}
    \nabla_{\theta} \ell_{\mathrm{RBC}}(\theta) \propto \nabla_{\theta} ( \delta^{2}\left(\mathcal{B}_{\mathrm{demo}}, 1\right) \nonumber \\ + \lambda_{\mathrm{samp}} \delta^{2}\left(\mathcal{B}_{\mathrm{samp}}, 0\right)+V\left(s_{0}\right))
\end{eqnarray}
In original paper, $ \lambda_{\mathrm{samp}} = 1$. 
Importantly, we can recognize that the SQIL agent sets the rewards of all demonstration data to 1 and the rewards of all sampling data to 0.

\section{Discriminator Soft Actor Critic} \label{secDSAC}

SQIL has shown that we can make our agent imitation efficiently by soft Q-learning with positive constant rewards to demonstration data. However, the method of determining this constant reward may not be good in some cases. 
For example, if the initial state $ s _ 0 $ given to the agent is significantly different from the domain of demonstrations, it becomes difficult to learn good behavior because the almost all rewards around it are zero.
The value is propagated when the agent continues to explore and happens to arrive near the state existed in demonstration data. However, it is also strongly affected by the discount rate $ \gamma $. 
Therefore, we propose Discriminator Soft Actor Critic (DSAC) to use a reward function $ R _ \varphi (s, a) $ parameterized by $ \varphi $ that gives a reward for the behavior is close to the demonstration data even in an unseen state. 
We show the overall of this algorithm in Figure~\ref{fig:overall}.

\subsection{Loss function}
We define the loss function as Equation~\ref{ll_DSAC} like Equation~\ref{RBC}.
\begin{equation}
    \ell_{\mathrm{DSAC}}(\theta) \triangleq \ell_{\mathrm{BC}}(\theta)+\lambda \delta^{2}\left(\mathcal{B}_{\mathrm{demo}} \cup \mathcal{B}_{\mathrm{samp}}, R _ \varphi \right)
    \label{ll_DSAC}
\end{equation}
In preparation, we separate the soft Bellman error of demonstrations and samples.
\begin{align}
    \ell_{\mathrm{DSAC}}(\theta) \triangleq \ell_{\mathrm{BC}}(\theta) & + \lambda_{\mathrm{demo}} \delta^{2}\left(\mathcal{B}_{\mathrm{demo}} , R _ \varphi \right) \nonumber \\ &+ \lambda_{\mathrm{samp}} \delta^{2}\left(\mathcal{B}_{\mathrm{samp}} , R _ \varphi \right)
    \label{l_DSAC}
\end{align}
We can rewrite it into a simple equation by deformation of formula for the gradient in Equation~\ref{l_DSAC}.
Here, we assume a continuous state space $ \mathcal{S} $ and a continuous action space $ \mathcal{A} $. 
In addition, we assume the network parameter of reward function $ \varphi $ is independent of the network parameter of actor critic $ \theta $. 
When focusing on the loss related to demonstration data, the gradient is rewritten by Equation~\ref{lossBC}. 
    \begin{align}
        & \nabla _ \theta \ell _{\mathrm{DSAC}_{\mathrm{demo}}} (\theta) \nonumber \\ 
        & = \sum _ {(s _ t, a _ t) \in \mathcal{B} _ {\mathrm{demo}}} - \left(\frac{ \nabla _ \theta Q_\theta(s_t, a_t) - \nabla _ \theta V _ \theta (s_t)}{\alpha} \right) \nonumber \\
        & + \lambda_{\mathrm{demo}} \delta^{2}\left(\mathcal{B}_{\mathrm{demo}} , R _ \varphi \right) 
        \label{nablaDSACdemo1}
    \end{align}
We substitute Equation~\ref{softBellman} for Equation~\ref{nablaDSACdemo1}.
\begin{align}
    & (\ref{nablaDSACdemo1}) = \sum _ {(s _ t, a _ t) \in \mathcal{B} _ {\mathrm{demo}}} - \left(\frac{ \nabla _ \theta Q_\theta(s_t, a_t) - \nabla _ \theta V _ \theta (s_t)}{\alpha} \right) \nonumber \\
    & + \lambda _{\mathrm{demo}} \sum _ {(s _ t, a _ t, s_{t+1}) \in \mathcal{B} _ {\mathrm{demo}}} \nabla _ \theta (Q_\theta(s_t, a_t) \nonumber \\ 
    & - \mathcal{R} _ \varphi (s_t, a_t) - \gamma V _ \theta (s_{t+1})) ^ 2 \nonumber \\ 
    & = \sum _ {(s _ t, s _ {t+1} ) \in \mathcal{B} _ {\mathrm{demo}}} \left(\frac{ \nabla _ \theta V_\theta(s_t) - \gamma \nabla _ \theta V _ \theta (s_{t+1})}{\alpha} \right) \nonumber \\
    & + \lambda _ {\mathrm{demo}}  \nabla _ \theta \delta ^ 2 \left( \mathcal{B} _ {\mathrm{demo}}, R _ \varphi + \frac{1}{2 \alpha \lambda_{\mathrm{demo}}} \right) \nonumber \\
    & \propto \sum _ {(s _ t, s _ {t+1} ) \in \mathcal{B} _ {\mathrm{demo}}} \nabla _ \theta \left(V_\theta(s_t) - \gamma V _ \theta (s_{t+1}) \right) \nonumber \\ 
    & + \alpha \lambda _ {\mathrm{demo}}  \nabla _ \theta \delta ^ 2 \left( \mathcal{B} _ {\mathrm{demo}}, R _ \varphi + \frac{1}{2 \alpha \lambda_{\mathrm{demo}}} \right)
    \label{nablaDSACdemo2}
\end{align}
Naturally, the soft value function $ V_\theta(s_t) = \mathbb{E} [ R _ \varphi (s_t, a_t) + \alpha \mathcal{H} (\pi_{\theta} ( \cdot | s_t)) ] + \gamma V(s_{t+1}) $ is established. Since we assume that $ \varphi $ is independent of $ \theta $, $ \nabla _ \theta \left(V_\theta(s_t) - \gamma V _ \theta (s_{t+1}) \right) = \nabla _ \theta (\alpha \mathcal{H} (\pi_{\theta}( \cdot | s_t))$. 

Here, SAC adjusts the entropy term so as to be greater than or equal to a hyperparameter constant value $\tilde {\mathcal{H}} $ by Equation~\ref{min_temp}. 
Therefore, we don't minimize $ \mathcal{H} (\pi_{\theta}) $ by the parameter $ \theta $.
As a result, we can represent the gradient with a simple formula.
\begin{equation} 
    (\ref{nablaDSACdemo2}) \propto \lambda _ {\mathrm{demo}}  \nabla _ \theta \delta ^ 2 \left( \mathcal{B} _ {\mathrm{demo}}, R _ \varphi + \frac{1}{2 \alpha \lambda_{\mathrm{demo}}} \right)
\end{equation}
Therefore, the gradient in Equation~\ref{l_DSAC} is also simple formula.
\begin{align}
    \nabla _ \theta \ell_{\mathrm{DSAC}}(\theta) & \propto \lambda_{\mathrm{demo}} \delta^{2}\left(\mathcal{B}_{\mathrm{demo}} , R _ \varphi + \frac{1}{2 \alpha \lambda_{\mathrm{demo}}} \right) \nonumber \\ 
    & + \lambda_{\mathrm{samp}} \delta^{2}\left(\mathcal{B}_{\mathrm{samp}} , R _ \varphi \right)
    \label{DSAC}
\end{align}
In practice, we found that the bonus term of the demonstration $ (2 \alpha \lambda_{\mathrm{demo}})^{-1} $ should be very small value for stable training.

\subsection{Reward function}
We suggested using a reward function parameterized by $ \varphi $ in Section~\ref{secDSAC}. 
In order to perform more robust and stable training, we adopt a binary classifier based on adversarial learning based on AIRL. 
We use the classifier $ D _ \varphi \left(s, a \right) $ to minimize the loss function in Equation~\ref{lossDSACreward}.
\begin{align}
    \ell (\varphi) \triangleq & - \mathbb{E}_{(s, a) \sim \mathcal{B} _ {\mathrm{demo}}}\left[\log \left(D_{\varphi}\left(s, a \right)\right)\right] \nonumber \\
    & -\mathbb{E}_{(s, a) \sim \mathcal{B} _ {\mathrm{samp}}}\left[ \log \left(1-D_{\varphi}\left(s, a \right)\right)\right]  \label{lossDSACreward}
\end{align} 
We define the reward function as Equation~\ref{DSACreward} with it.
\begin{equation}
    \mathcal{R} _ \varphi \left(s, a \right) = \log \left(D_{\varphi}\left(s, a \right)\right) - \log \left(1-D_{\varphi}\left(s, a \right)\right)  \label{DSACreward}
\end{equation}
When the classifier determines that the pair of state and action is in $ \mathcal{B} _ {\mathrm{demo}} $, the reward is positive. In contrast, when it determines the pair is in $ \mathcal{B} _ {\mathrm{samp}} $, the reward is negative.

In our experiments, we adopt GAN discriminator with zero-centered gradient penalty~\cite{thanh2019improving}. 
\begin{figure*}[htpb]
    \centering
      \begin{tabular}{c}
   
        \begin{minipage}{0.50\hsize}
          \centering
            \includegraphics[width=0.85\hsize]{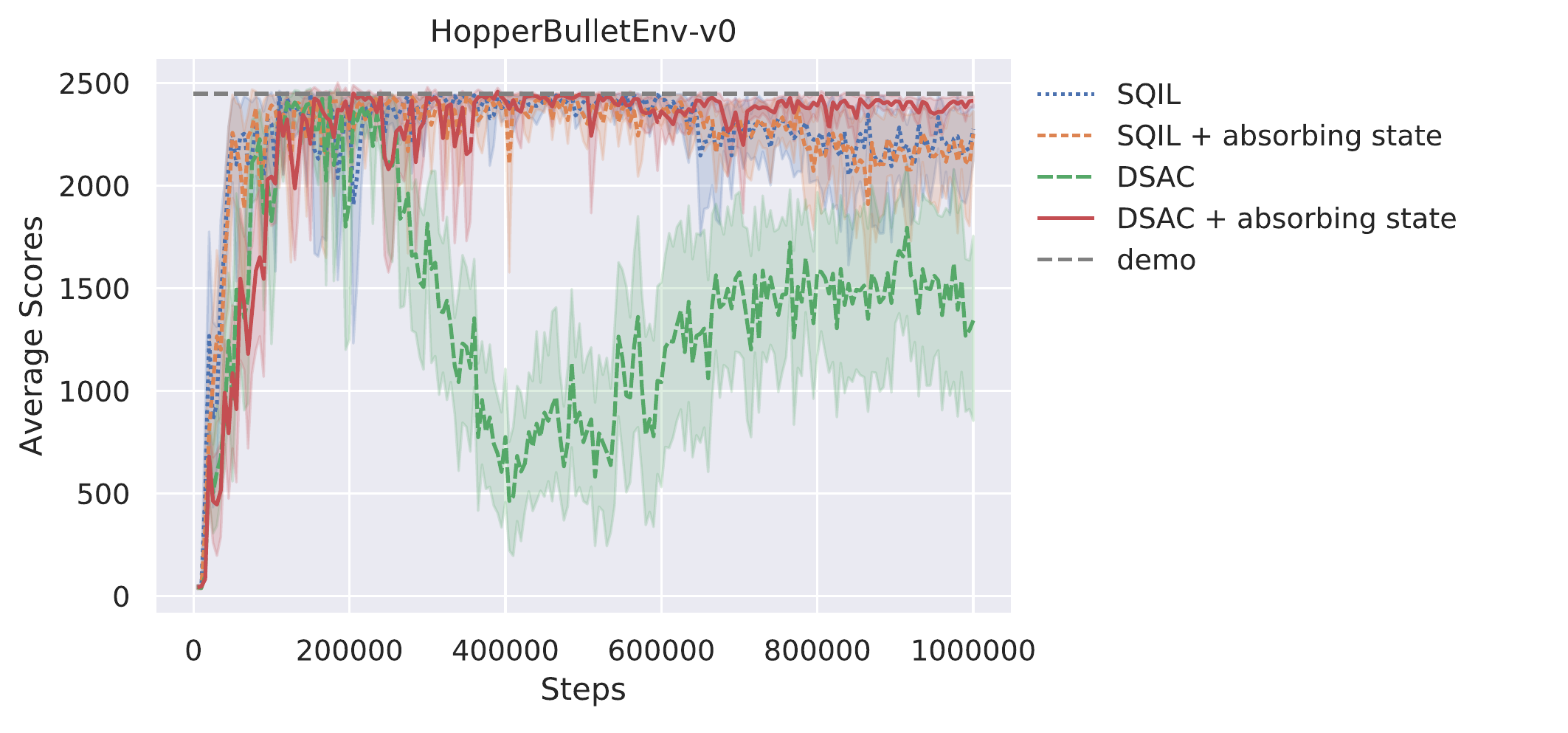}
        \end{minipage}
   
        \begin{minipage}{0.50\hsize}
          \centering
            \includegraphics[width=0.85\hsize]{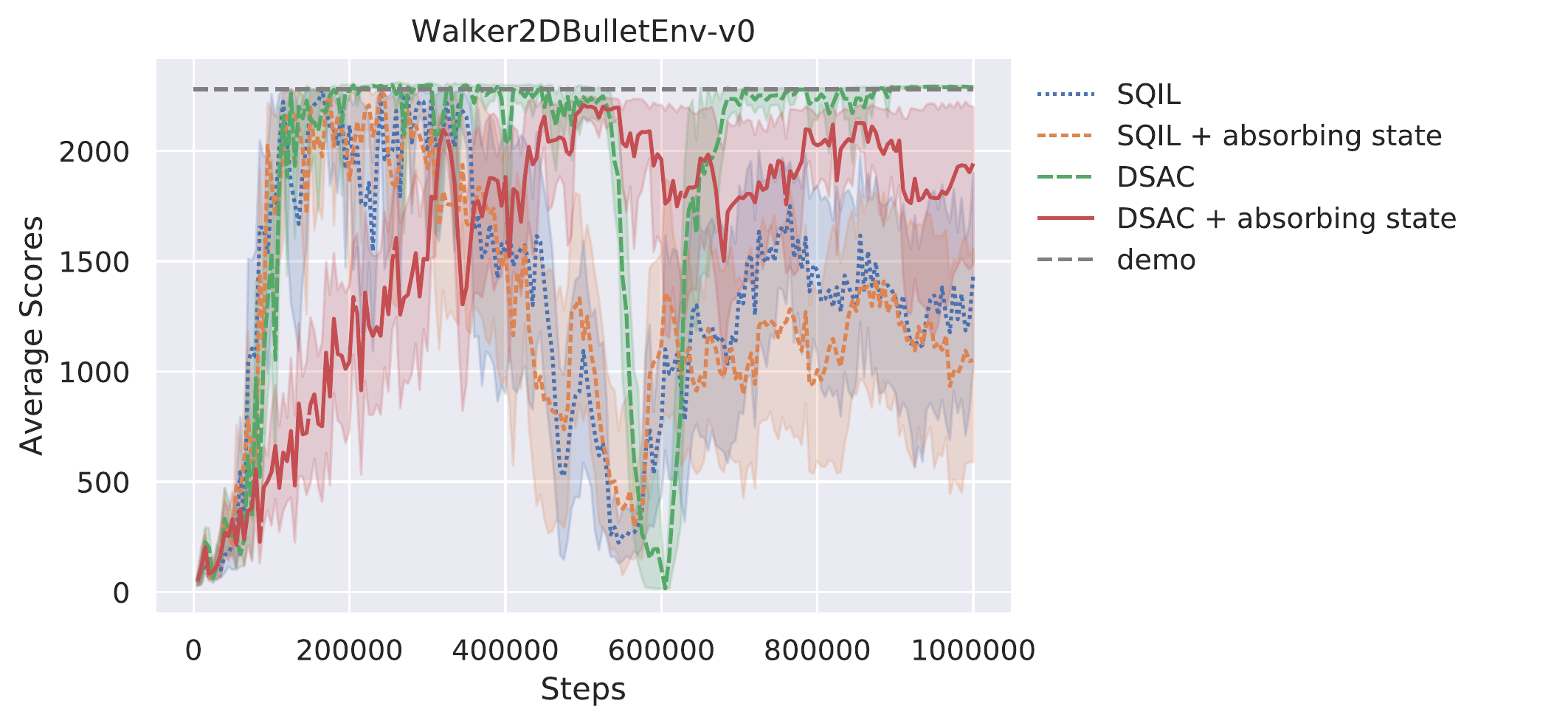}
        \end{minipage} \\
   
        \begin{minipage}{0.50\hsize}
            \centering
              \includegraphics[width=0.85\hsize]{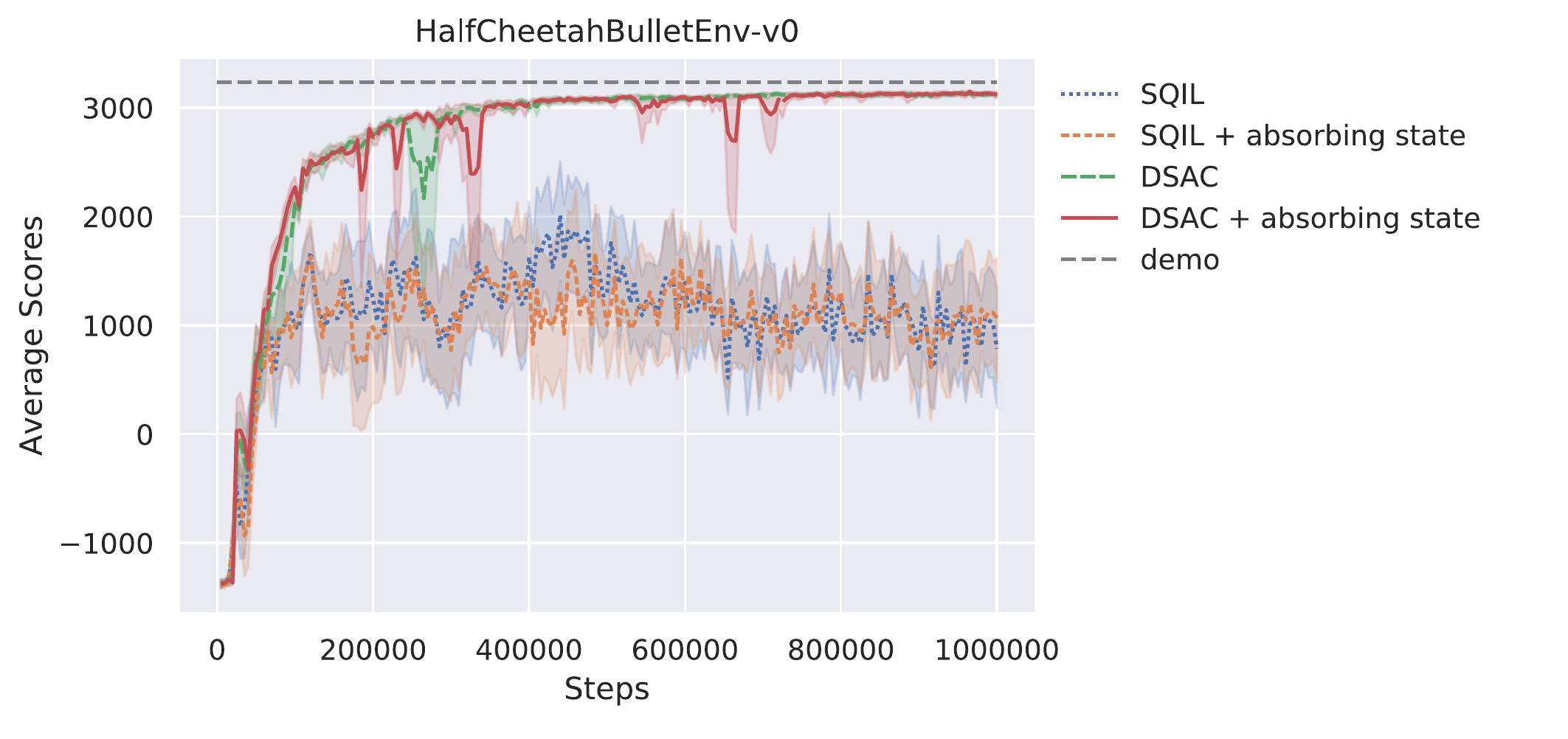}
        \end{minipage}
   
        \begin{minipage}{0.50\hsize}
          \centering
            \includegraphics[width=0.85\hsize]{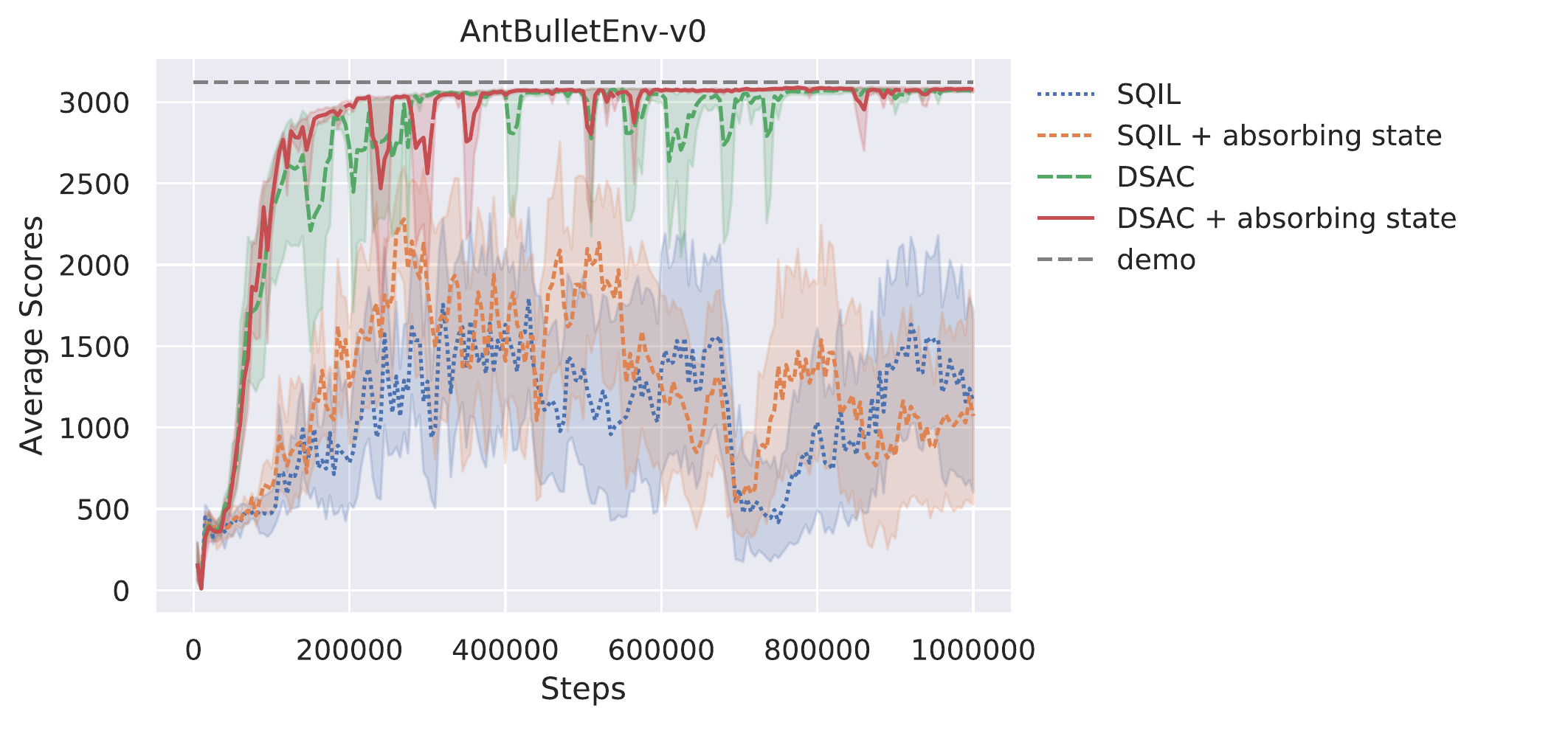}
        \end{minipage}
   
      \end{tabular}
    \caption{Comparisons of scores using 4 expert demonstrations.}
    \label{fig:scores}
\end{figure*} 
\subsection{Reducing the bias of reward function}
As reported by Kostrikov et al.~\shortcite{kostrikov2018discriminator}, DSAC may need to address the bias of the reward function. 
Therefore, it also needs to explicitly learn the value of the absorption state. 
Specifically, we apply an absorbing state wrapper. 
We show the algorithm in Algorithm~\ref{alg:dsac}.

\begin{algorithm}[htbp]  
    \caption{Discriminator Soft Actor Critic with absorbing state wrapper}         
    \label{alg:dsac}
    \begin{algorithmic}[1]
    \State Require a replay buffer of demonstaration data $ \mathcal{B}_{\mathrm{demo}} $ 
    \State Initialize parameters $ \theta $ and Initialize an empty replay buffer of the agent $ \mathcal{B}_{\mathrm{samp}} \leftarrow \emptyset $
    \Procedure{WrapForAbsorbingStates}{$ \tau $}
        \If {$ s_T $ is a terminal state not caused by time limits}
            \State $ \tau \leftarrow \tau \setminus \{ (s_T, a_T, \cdot, s'_{T}) \} \cup \{ (s_T, a_T, \cdot, s_a ) \} $
            \State $ \tau \leftarrow \tau \cup \{ (s_a, \cdot, \cdot, s_a )\} $
        \EndIf
        \State \Return $ \tau $
    \EndProcedure
    \For {$ \tau = \{ (s_t, a_t, \cdot, s'_t)\}^T_{t=1} \in \mathcal{B}_{\mathrm{demo}} $ }
    \State $ \tau \leftarrow WrapForAbsorbingStates (\tau) $ 
    \EndFor
    \For {$ n = 1, 2, \dots $}
        \State Sample $ \tau = \{ (s_t, a_t, \cdot, s'_t)\}^T_{t=1} $ with $ \pi_\theta $
        \State $ \mathcal{B}_{\mathrm{samp}} \leftarrow \mathcal{B}_{\mathrm{samp}} \cup WrapForAbsorbingStates (\tau)$ 
        \For {$ i = 1, 2, \dots, | \tau | $}
            \State $ \{ (s_t, a_t, \cdot, \cdot)\}^M_{t=1} \sim \mathcal{B}_{\mathrm{demo}} $
            \State $ \{ (s'_t, a'_t, \cdot, \cdot)\}^M_{t=1}\sim \mathcal{B}_{\mathrm{samp}} $ 
            \State Calcurate the loss of $ D $ \Comment {See Equation~\ref{lossDSACreward}}
            \State Update $ D $ with GAN + zero-centered GP
        \EndFor
        \For {$ i = 1, 2, \dots, | \tau | $}
            \State $ \mathcal{M} _ {\mathrm{demo}} = \{ (s_t, a_t, \cdot, s'_t)\}^{M_{\mathrm{demo}}}_{t=1} \sim  \mathcal{B}_{\mathrm{demo}}$ 
            \State $ \mathcal{M} _ {\mathrm{samp}} = \{ (s_t, a_t, \cdot, s'_t)\}^{M_{\mathrm{samp}}}_{t=1} \sim  \mathcal{B}_{\mathrm{samp}}$ 
            \For {$ m = 1, 2, \dots, M_\mathrm{demo} $}
                \State $ r_{\mathrm{demo}_m} \leftarrow R(s_m, a_m)$ \Comment {See Equation~\ref{DSACreward}}
            \EndFor
            \For {$ m = 1, 2, \dots, M_\mathrm{samp} $}
                \State $ r_{\mathrm{samp}_b} \leftarrow R(s_m, a_m)$ \Comment {See Equation~\ref{DSACreward}}
            \EndFor
            
            \State Update $ \theta $ with SAC \Comment {See Equation~\ref{DSAC}}
        \EndFor
    \EndFor
    \end{algorithmic}
\end{algorithm}

\subsection{Relation to existing methods}
The loss functions of SQIL and DAC can be regarded as special cases of DSAC. 
For example, if the reward function $ R (s, a) $ always returns 0 and the bonus term $ (2 \alpha \lambda_{\mathrm{demo}})^{-1} $ is 1 in Equation~\ref{DSAC}, it is the same as the loss function of SQIL.
Also, if we set the bonus term to 0 and we use TD3~\cite{fujimoto2018addressing} instead of SAC as a reinforcement learning algorithm, it is the same as the loss function of DAC.

\section{Experimental Evaluation}

Our experiments aim to compare DSAC to SQIL with a few demonstration data. 
We evaluate it on PyBullet~\cite{coumans2016pybullet}.
It has reimplemented MuJoCo~\cite{todorov2012mujoco} in which is popular benchmarks for continuous control simulated.

For the critic and policy networks we used the same architecture as in~\cite{haarnoja2018soft2}: a 2 layer MLP with ReLU activations and 256 hidden units. We also add gradient clipping to the actor network with clipping value 40 similar to~\cite{kostrikov2018discriminator}.
For the discriminator we used the same architecture as in~\cite{ho2016generative}: a 2 layer MLP with 100 hidden units and tanh activations. We trained all networks with the Adam optimizer~\cite{kingma2014adam} and decay learning rate by starting with initial learning rate of $ 10 ^{-3} $ and decaying it by 0.5 every $ 10^5 $ training steps for the actor network.
Following Fujimoto et al.~\shortcite{fujimoto2018addressing} and Kostrikov~\shortcite{kostrikov2018discriminator}, we perform evaluation using 10 different random seeds. 
For each seed, we compute average episode reward using 10 episodes and
running the policy without random noise.
In order to prepare demonstration, we store 4 trajectories in $ \mathcal{B} _ {\mathrm{demo}} $ after the agent trains by SAC algorithms for 3 million steps. 
We implemented DSAC and an expert agent using ChainerRL~\cite{fujita2019chainerrl}.
\footnote{Our code is available at \url{https://github.com/dnishio/DSAC}.}

We show other hyperparameters in Table~\ref{tab:params}.

\begin{table}[htbp]
    \caption{DSAC hyperparameters} 
    \label{tab:params}
    \centering
    \begin{tabular}{|l|l|}
        \hline
        Parameter & Value \\
        \hline
        discount rate $ (\gamma) $ & 0.99 \\
        maximum size of a replay buffer $ (\mathcal{B}_{\mathrm{samp}}) $ & $ 5 \cdot 10^5 $ \\
        entropy target $ (\tilde{\mathcal{H}}) $ & $ - \mathrm{dim} (\mathcal{A}) $  \\
        target smoothing coefficient & $ 5 \cdot 10^{-3} $ \\
        size of minibatch $ (M)$ & 100 \\
        number of training steps & 1,000,000 \\
        warm-up & 10,000 \\
        interval of evaluation  &  per 5,000 \\
        seed of demonstaration & 0 \\
        seeds of SQIL and DSAC & $\{1, 2, ..., 10\}$ \\
        \hline
    \end{tabular}

\end{table}

For each task, we compare the following four algorithms:
\begin{enumerate}
    \item SQIL (based on SAC)~\cite{reddy2019sqil}
    \item SQIL (based on SAC) + absorbing state wrapper
    \item DSAC 
    \item DSAC + absorbing state wrapper
\end{enumerate}

\subsection{Comparing the scores}

Evaluation results of the DSAC algorithm are shown in Figure~\ref{fig:scores} as are the SQIL results.
We plotted the scores of demonstration by computing average score of 100 episodes by the expert agent.
In Hopper and Walker2D environments, the performance of DSAC drops significantly, but it can be improved by using the absorbing state wrapper. 
Moreover, it outperforms SQIL's performance in all environments. 
The disadvantage of DSAC may imitate slower than SQIL for simple tasks because it requires learning a reward function. 
Whether we choose DSAC or SQIL is dependent on the difficulty of tasks. 
For example, in Walker2D, we can obtain a good performance by early stopping~\cite{yao2007early} in about 200,000 steps. 
However, for complex tasks, the performance may drop before imitating completely.
 
\subsection{Comparing the rewards for soft Bellman error}

\begin{figure*}[htpb]
    \centering
      \begin{tabular}{c}
   
        \begin{minipage}{0.50\hsize}
          \centering
            \includegraphics[width=0.85\hsize]{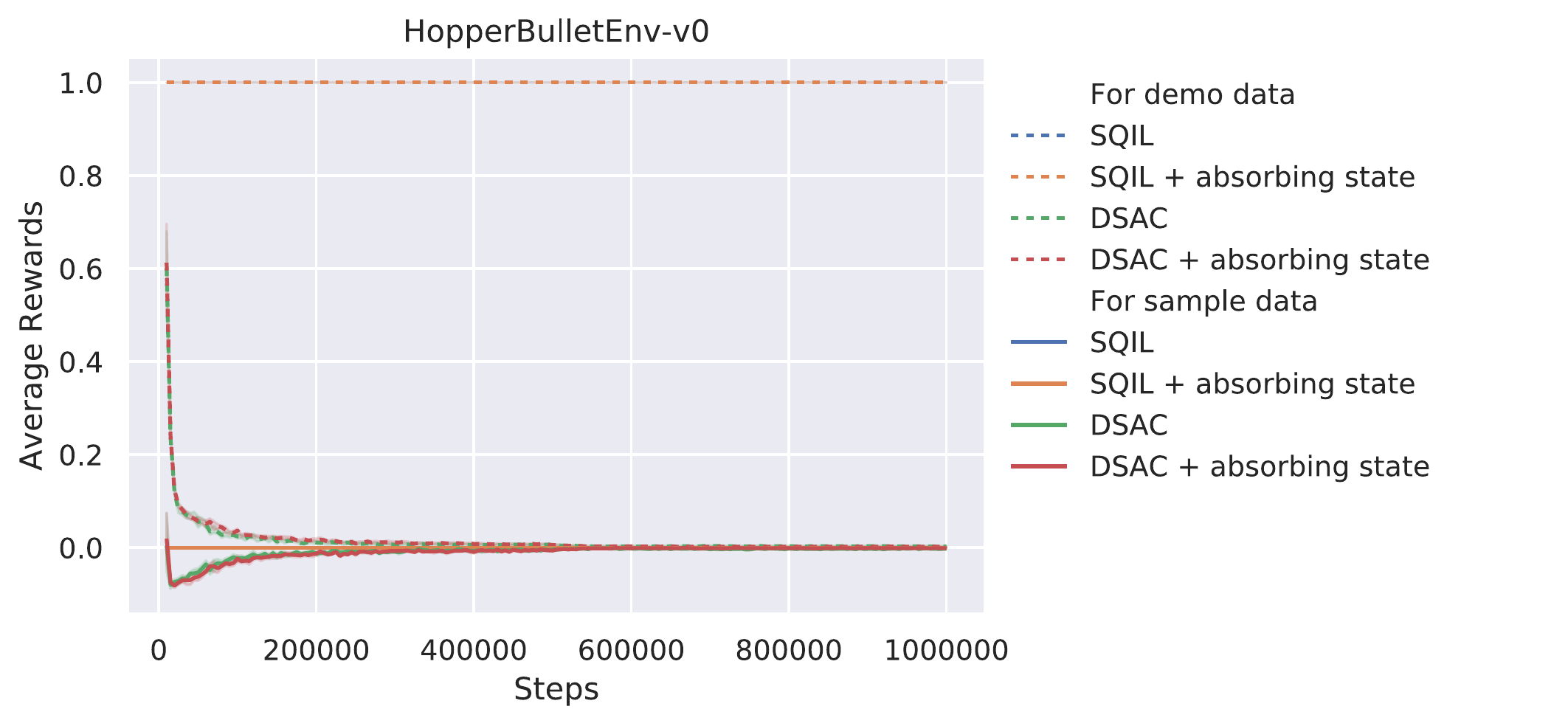}
        \end{minipage}
   
        \begin{minipage}{0.50\hsize}
          \centering
            \includegraphics[width=0.85\hsize]{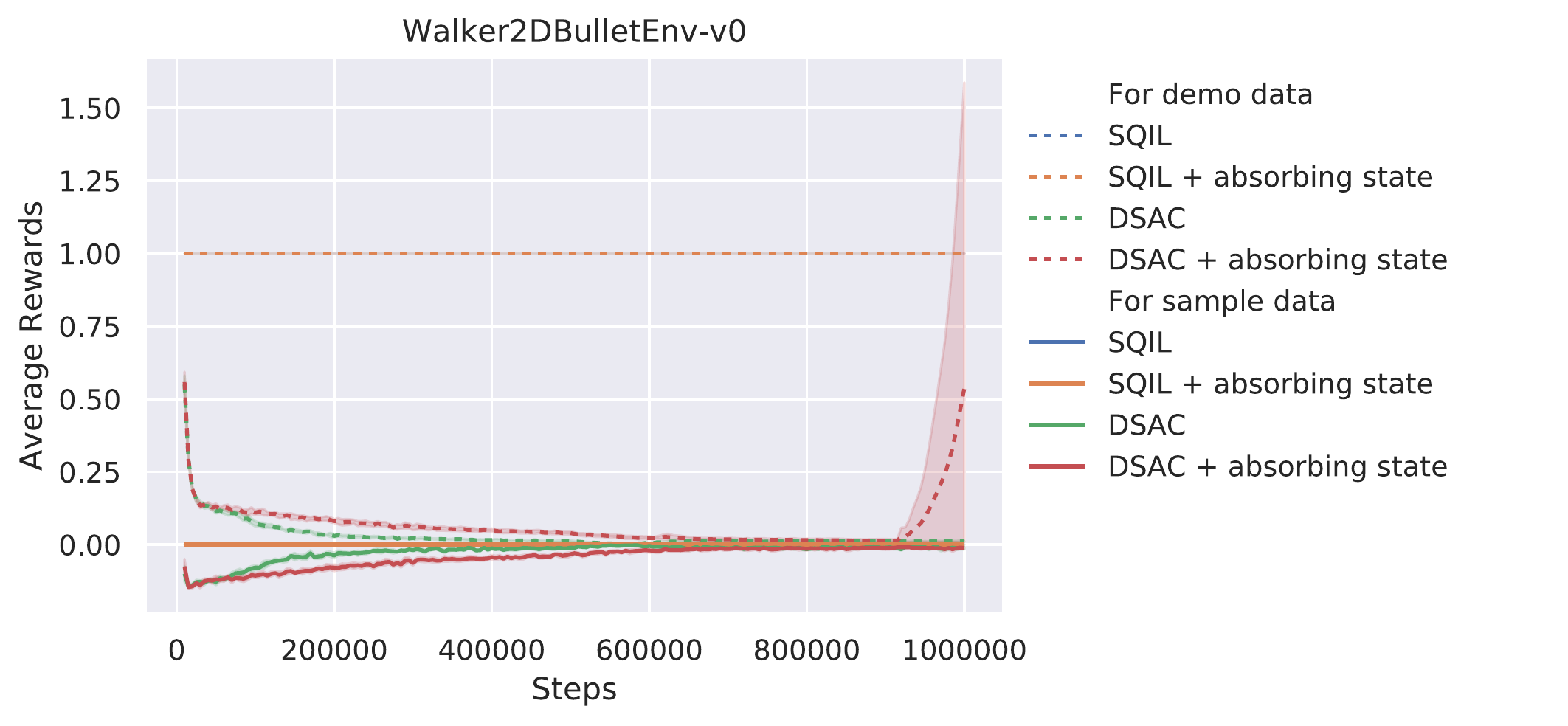}
        \end{minipage} \\
   
        \begin{minipage}{0.50\hsize}
            \centering
              \includegraphics[width=0.85\hsize]{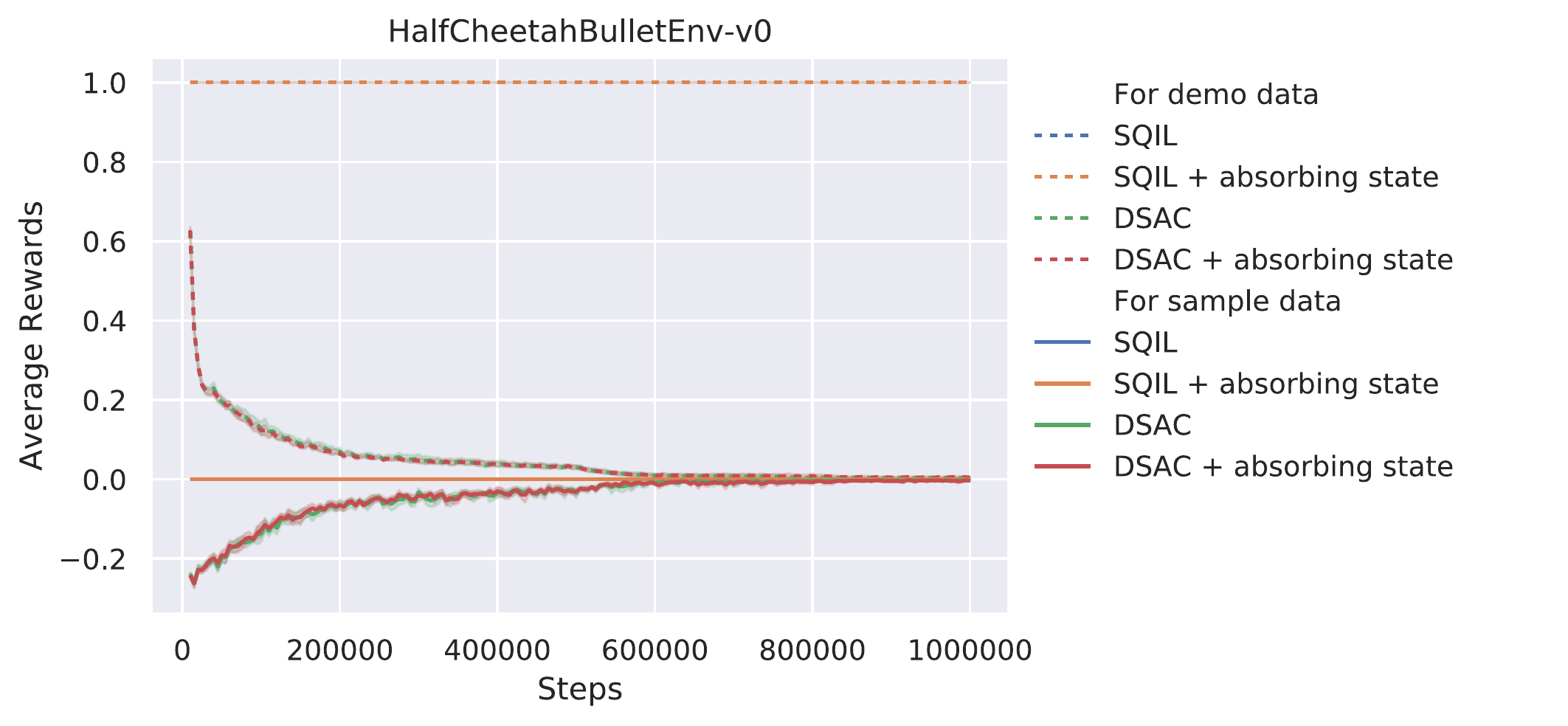}
        \end{minipage}
   
        \begin{minipage}{0.50\hsize}
          \centering
            \includegraphics[width=0.85\hsize]{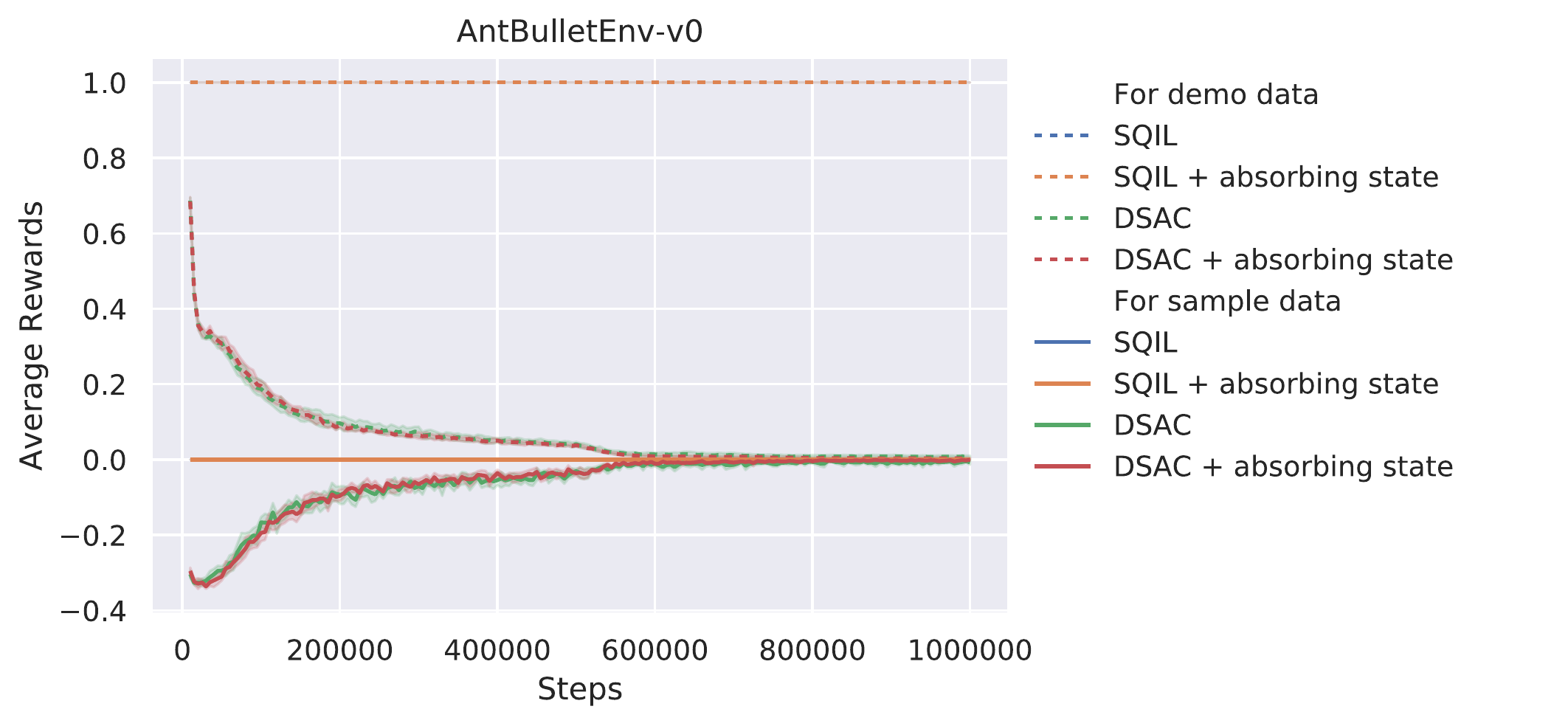}
        \end{minipage}
   
      \end{tabular}
    \caption{Comparisons of rewards of demonstaration and sampling data.}
    \label{fig:rewards}
\end{figure*}  
According to Equation~\ref{DSACreward}, when an agent imitates demonstration completely, the reward function returns zero. 
Therefore, if the discriminator is able to learn sufficiently, it is possible to confirm how much it can imitate. 
Comparing the rewards of DSAC and SQIL in Figure~\ref{fig:rewards}, SQIL learns with a constant reward, so even if it becomes possible to imitate, the rewards are not the same. 
On the other hand, DSAC gives the reward for the demonstration is large in the early stages of learning because it has not been able to imitate yet. 
Then, as it can imitate, the rewards are gradually approaching zero.

\section{Conclusion}
In this paper, we proposed Discriminator Soft Actor Critic (DSAC) as a robust and data-efficient imitation learning method. 
In contrast to the conventional method SQIL, we showed that the reward function helped to give more detail rewards for the pair of state and action instead of constant rewards. 
We evaluated on four experiments of PyBullet, and the performance was better than the current imitation learning. 
Furthermore, we showed that more stable learning could be achieved by learning the value of the absorbing state in order to stabilize the adversarial inverse reinforcement learning as DAC.
As future work, we should verify whether DSAC is valid in a high dimensional state space or action space. In addition, we should evaluate the effects of reward bonus terms on demonstrations.

\bibliographystyle{named}
\bibliography{ijcai20_dnishio}

\end{document}